\documentclass[sigconf,anonymous=false]{acmart}

\setcopyright{none}
\copyrightyear{2021}
\acmYear{2021}
\acmConference[CIKM '21]{CIKM '21: ACM International Conference on Information and Knowledge Management}{November 01--05, 2021}{Gold Coast, Queensland, Australia}
\acmBooktitle{CIKM '21: ACM International Conference on Information and Knowledge Management,
  November 01--05, 2021, Gold Coast, Queensland, Australia}

\usepackage[T1]{fontenc}

\usepackage[utf8]{inputenc}

\usepackage{microtype}

\usepackage{mathtools}
\usepackage{soul,color}
\usepackage{xcolor,colortbl}
\usepackage{hyperref}
\usepackage{amsmath}
\usepackage{makecell}
\usepackage{enumitem}
\usepackage{array}
\usepackage[capitalize]{cleveref}
\usepackage{multirow}
\usepackage[ruled,vlined]{algorithm2e}

\crefname{algocf}{Alg.}{Algs.}
\Crefname{algocf}{Algorithm}{Algorithms}

\usepackage{tikz}

\newcommand{\modelname}{\text{LEMPEx}}
\DeclareMathOperator{\TF}{\text{T5}}

\usepackage[font={footnotesize}]{caption}

\newcommand{\PreserveBackslash}[1]{\let\temp=\\#1\let\\=\temp}
\newcolumntype{C}[1]{>{\PreserveBackslash\centering}p{#1}}
\newcolumntype{R}[1]{>{\PreserveBackslash\raggedleft}p{#1}}
\newcolumntype{L}[1]{>{\PreserveBackslash\raggedright}p{#1}}

\definecolor{darkpastelgreen}{rgb}{0.01, 0.75, 0.24}
\definecolor{brilliantlavender}{rgb}{0.96, 0.73, 1.0}
\definecolor{Gray1}{rgb}{0.91,0.925, 0.937}
\definecolor{cambridgeblue}{rgb}{0.0, 0.8, 0.6}
\definecolor{Gray2}{rgb}{0.87, 0.886, 0.902}
\definecolor{Gray3}{rgb}{0.808, 0.831, 0.855}
\definecolor{Gray4}{rgb}{0.678,0.71, 0.741}
\definecolor{darkgreen}{rgb}{0.0, 0.5, 0.0}
\definecolor{almond}{rgb}{0.99, 0.87, 0.9}
\definecolor{ghostwhite}{rgb}{0.98, 0.81, 0.69}
\definecolor{Blue1}{rgb}{0.792, 0.941, 0.973}
\definecolor{Blue2}{rgb}{0.678, 0.91, 0.957}
\definecolor{Blue3}{rgb}{0.565, 0.878, 0.937}
\definecolor{Blue4}{rgb}{0.282, 0.749, 0.89}
\definecolor{Yellow1}{rgb}{1, 0.914, 0.306}
\definecolor{Yellow2}{rgb}{1, 0.886, 0.275}
\definecolor{Yellow3}{rgb}{1, 0.855, 0.239}

\definecolor{Green0}{rgb}{0.909, 0.992, 0.886}
\definecolor{Green1}{rgb}{0.843, 0.960, 0.839}
\definecolor{Green2}{rgb}{0.635, 0.854, 0.627}

\begin{document}

\title[LEMPEx: Learning to EMPathize using Exemplars]{Exemplars-guided Empathetic Response Generation Controlled by the Elements of Human Communication}

\author{Navonil Majumder$^\ast$}
\affiliation{%
\def\institution{}
\institution{Singapore University of Technology and Design}
 \country{Singapore}}
\email{navonil\_majumder@sutd.edu.sg}

\author{Deepanway Ghosal$^\ast$}
\affiliation{%
\def\institution{}
\institution{Singapore University of Technology and Design}
 \country{Singapore}}
\email{deepanway\_ghosal@mymail.sutd.edu.sg}

\author{Devamanyu Hazarika}
\affiliation{%
  \institution{National University of Singapore}
 \country{Singapore}}
 \email{hazarika@comp.nus.edu.sg}
 
 \author{Alexander Gelbukh}
\affiliation{%
 \institution{Instituto Polit\'ecnico Nacional}
 \city{Mexico City}
 \country{Mexico}}
\email{gelbukh@cic.ipn.mx}

\author{Rada Mihalcea}
\affiliation{%
 \institution{University of Michigan}
 \city{Ann Arbor}
 \state{Michigan}
 \country{USA}}
\email{mihalcea@umich.edu}

\author{Soujanya Poria}
\affiliation{%
\def\institution{} 
\institution{Singapore University of Technology and Design}
 \country{Singapore}}
 \email{sporia@sutd.edu.sg}
 
\def\shortauthors{Majumder, Ghosal, Hazarika, Gelbukh, Mihalcea, Poria}
 
\begin{abstract}
The majority of existing methods for empathetic response generation rely on the emotion of the context to generate empathetic responses. However, empathy is much more than generating responses with an appropriate emotion. It also often entails subtle expressions of understanding and personal resonance with the situation of the other interlocutor. Unfortunately, such qualities are difficult to quantify and the datasets lack the relevant annotations. To address this issue, in this paper we propose an approach that relies on exemplars to cue the generative model on fine stylistic properties that signal empathy to the interlocutor. To this end, we employ dense passage retrieval to extract relevant exemplary responses from the training set. Three elements of human communication---emotional presence, interpretation, and exploration---and sentiment are additionally introduced using synthetic labels to guide the generation towards empathy. The human evaluation is also extended by these elements of human communication. We empirically show that these approaches yield significant improvements in empathetic response quality in terms of both automated and human-evaluated metrics. The implementation is available at \url{https://github.com/declare-lab/exemplary-empathy}.
\end{abstract}


\maketitle

\section{Introduction} \label{sec:intro}

Empathy is a very human characteristic often defined as the ability to understand and emotionally comprehend the feelings of others intimately. An entire field of research has grown around empathy that delves into its fundamentals~\cite{Singer09The}. Progress has been made into understanding empathy from the angle of affect~\cite{Smith06Cognitive}, personality traits, and demographics~\cite{Dymond50Personality,Eisenberg14,Krebs75Empathy}. 
Understanding of empathy has widespread applications
in healthcare,
such as in psychotherapy~\cite{Bohart97Empathy} and
in general care improvement~\cite{Mercer02Empathy}, among others.

\begin{figure}[ht]
    \centering
    \includegraphics[width=\linewidth]{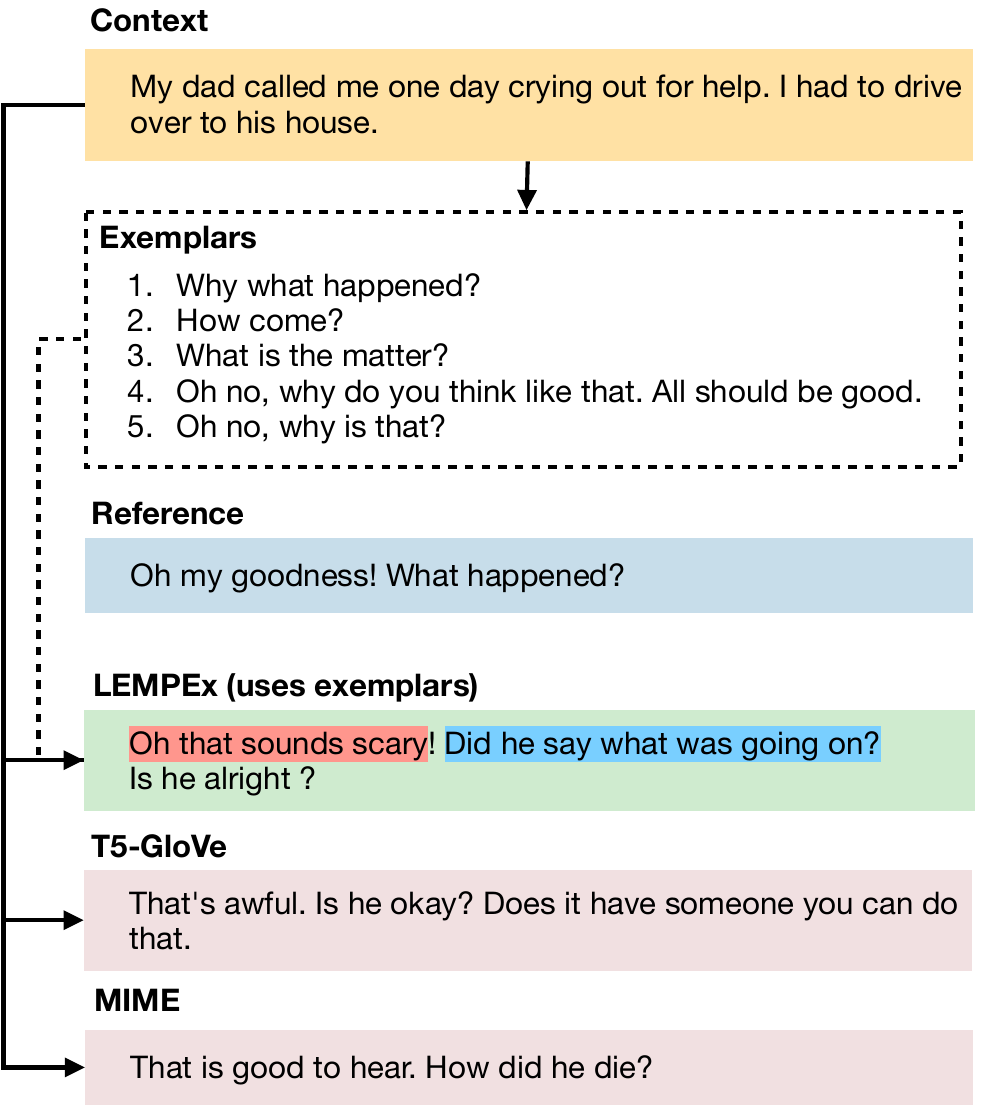}
    \caption{\footnotesize Influence of exemplars on the response.}
    \label{fig:teaser}
\end{figure}

Owing to its intrinsic influence on conversations, empathy has garnered much interest in the NLP community. Recent works on empathetic response generation~\cite{lin-etal-2019-moel,rashkin-etal-2019-towards,majumder-etal-2020-mime,li-etal-2020-empdg} singularly focus on establishing a relation between emotion of the contextual utterances and the response. Although response emotion is a key element of empathy, it is far from the complete picture. In fact, the very definition of empathy, which involves visceral understanding of the feeling of other people, is quite subjective and vague
and, as such,
is quite unwieldy for the empathetic response generation task.

To sidestep this issue, we adopt the idea of exemplars~\cite{chen-etal-2019-controllable} for controllable text generation, where a set of sample responses from the training set, semantically related to the input context, 
is retrieved (%
using fine-tuned dense passage retrieval (DPR)~\cite{karpukhin-etal-2020-dense} model) and fed to the response generator as templates~\cite{DBLP:conf/ijcai/CaiCSZY20}. These template responses guide the generator with stylistic and thematic cues on the response that are perceived as empathetic to the user. This is similar to providing hints for solving a difficult mathematical problem to significantly narrow down the search space.
Similarly, we surmise that the exemplars restrict the search space for the decoder. 

Our model is called \modelname{}: Learning to EMPathize using Exemplars. \cref{fig:teaser} illustrates an instance where the support from exemplars leads to an empathetically resonant and thematically rich response from our model, \modelname{}. 
The phrase ``\emph{Oh that sounds scary!}'' is emotionally appropriate to communicate strong empathy. Moreover, the response makes a thematically
appropriate
query with ``\emph{Did he say what was going on?}''. Based on the similarity between the extracted exemplars and the generated response, we posit that the exemplars helped \modelname{} regulate the nature of the query and emotional intensity. The \emph{T5-GloVe} model,
on the other hand, does not use exemplars. It is probably this what 
leads it
to a partly generic and partly unintelligible query: ``\emph{Is he okay? Does it have someone you can do that.}''.  ``\emph{That's awful}'' phrase is not quite empathetic in this situation; rather, it might be construed as an overreaction. Finally, \emph{MIME} also misinterprets the situation and asks an inappropriate question. We delve deeper into this in \cref{sec:DPR_impact}.

To further regulate the response generation, we introduce four auxiliary losses that quantify four important aspects related to empathy: \emph{emotional presence}, \emph{interpretation}, \emph{exploration}, and \emph{sentiment}. Many research works~\cite{sharma2020empathy,sharma2021facilitating,watson_client-centered_2002} identify the former three as key aspects of empathy and human communication in general. The role of adequate \emph{emotional presence} is to form an emotional bond with the user. In \cref{fig:teaser}, the response phrase ``\emph{Oh that sounds scary!}'' is meant to achieve this. Then, correct \emph{interpretation} is key in regulating emotion and generating context-appropriate comments and queries that exude empathy. \emph{Exploration} through relevant and appropriate queries signals 
understanding and interest of the empathetic agent to the user, which
is
perceived as empathy by the user. The phrase ``\emph{Did he say what was going on? Is he alright?}'' is meant to communicate these two attributes. Finally, inspired by the idea of emotion grouping---these groups essentially represent positive and negative sentiment---by \citet{majumder-etal-2020-mime}, we also add sentiment as an extension of \emph{emotional presence}.

To this end, we generate synthetic labels for \emph{emotional presence}, \emph{interpretation}, and \emph{exploration} attributes for the gold responses by training classifiers for them with the \textsc{EmpathyMentalHealth}~\cite{sharma2020empathy} dataset. For \emph{sentiment}, we get scores from VADER~\cite{vader}. Subsequently, these labels are predicted from the decoder hidden states to prime the decoder to generate responses with these four attributes at a suitable level. These attributes may be viewed as an external signal to guide the generation process.

To gauge the empathetic quality of the generated responses at a deeper level, we also obtain human evaluation on \emph{emotional presence}, \emph{interpretation}, and \emph{exploration} attributes. This gives us better sense of the generation quality at a more fundamental level. To the best of our knowledge, we are the first to delve into such metrics to assess empathy in generated responses.

Evaluation of empathetic responses is reliant on human evaluation which is resource consuming. To ameliorate this, we repurpose the aforementioned synthetic-label predictors for the empathy-related attributes \emph{emotional presence}, \emph{interpretation}, \emph{exploration}, and \emph{sentiment} as automatic evaluators on corresponding attributes. The generated response is fed to the these predictors and the resultant label is compared to the synthetic label of the gold response to obtain macro-F1 and mean absolute error (MAE) scores. In our experiments, we observed good correlation with the other human evaluated metrics across all the discussed models that suggests the reliability of these automated metrics.

Later on in this paper, we empirically show that the combination of these approaches yields strong empathetic responses, outshining the state-of-the-art EmpDG~\cite{li-etal-2020-empdg} and other baselines. 

In a nutshell, the contribution of this paper is three-fold:
\begin{enumerate}
    \item We mine exemplary responses from the training set, using DPR, to guide the response generation;
    \item Empathy-inducing attributes, namely, \emph{emotional presence}, \emph{interpretation}, \emph{exploration}, and \emph{sentiment}, are introduced as regularizer losses by the means of synthetic labels.
    \item We conduct a thorough human and automatic evaluation based on the 
    these
    attributes, 
    for
    better understanding of the empathy generation capability of models.
\end{enumerate}

\section{Background}

\paragraph
{\textbf{Text-To-Text-Transfer-Transformer (T5)}}
Our 
model is based on the popular T5 model. The Text-to-text-transfer-transformer (T5) is an encoder-decoder based sequence-to-sequence transformer model \cite{raffel2020exploring}. In the T5 framework, every natural language processing task, including question answering, language translation, text classification, is framed as a sequence-to-sequence text generation problem. This formulation enables the use of identical architecture and loss functions without any changes across a diverse set of tasks. 

T5 follows the self-attention-based encoder-decoder transformer \cite{vaswani2017attention} with slight modifications in the architecture and a different position embedding strategy. As opposed to using a fixed sinusoidal position embedding as the original transformer model, T5 uses relative position embeddings that are learned based on the offset between key and query vector. This relative position embedding has been found to work better in practice \cite{shaw2018self,huang2018music}. 
T5 is pre-trained on the Colossal Clean Crawled Corpus (C4) dataset. A modified masked language modeling (MLM) objective function is used to pre-train the model.

\paragraph
{\textbf{Retrieval-Based Generation}}
Typically, generative models in NLP employ auto-regressive decoders to generate sentences. However, recent works have shown that simply relying on the model's parameters might be insufficient for specific tasks and might also lead towards undesired effects like hallucinations~\cite{DBLP:journals/corr/abs-2104-07567}. In particular, within the literature of controlled text generation, there is increasing focus to provide these models with additional support either through non-parametric access to retrieved content as inputs or through auxiliary losses. \cite{DBLP:journals/tacl/GuuHOL18} proposed a novel strategy that performed language modeling using retrieving fluent responses from human sentences in training data and then editing them. Inspired by this, \citet{DBLP:conf/emnlp/WestonDM18} proposed a retrieve-then-refine framework that involved retrieving relevant responses for a dialog context and then appending these retrievals along with the dialog context to train the overall generator. This simple idea demonstrated good performance to the otherwise daunting task of response generation. Similar approaches were also explored in  \cite{DBLP:conf/ijcai/CaiCSZY20} where the retrieval process was designed to ensure both textual similarity and topical match with the generated response.

While these approaches have been studied in generative applications like language modeling, open-domain dialog, or style transfer, to the best of our knowledge, we are the first to perform a systematic study of this paradigm for empathetic response generation. As we show in the results, our exemplar-guided generation for empathy becomes a better empathetic generator and a simpler solution than previous works in this application.

\paragraph
{\textbf{Related Works in Text Generation}}
Lately, significant strides have been made in open-domain conversational models~\cite{serban2016generative, vinyals2015neural, wolf2019transfertransfo}. 
Models for generating persona-consistent~\cite{zhang-etal-2018-personalizing}
and varied~\cite{cai2018skeletontoresponse}
responses have been developed. 
Their responses
are not necessarily empathetic.
One aspect of generating such responses involves regulating emotions and sentiments \cite{fung-etal-2016-zara-supergirl,Winata2017NoraTE,bertero-etal-2016-real}. 
GAN-based~\cite{goodfellow2014generative} model for emotion-specific generation has been proposed by \citet{ijcai2018-618}. \citet{zhou-wang-2018-mojitalk} consider emojis as emotion labels. They have also proposed an attention-based~\cite{luong2015effective} Seq-to-Seq~\cite{sutskever2014sequence} model with Conditional Variational Autoencoder~\cite{sohn2015learning} for emotion-specific response generation. 
They, however, require emotion prompt for response generation which may not necessarily be empathetic.
Similarly, \citet{wu2019simple}'s dual-decoder network responds with input sentiment. 
\citet{inproceedings} devise a reinforcement learning formulation that maximizes the user's sentiment on the generated response. Only recently, following the construction of \texttt{EmpatheticDialogues}~\cite{rashkin-etal-2019-towards} dataset, 
work~\cite{lin-etal-2019-moel,li-etal-2020-empdg,majumder-etal-2020-mime} has been done on emotion prompt-free automated empathetic response generation. However, these works heavily 
rely
on the emotion labels of the context, which does not capture the entire picture of empathy. 
Therefore, we leverage exemplar-based generation that has been quite successful in many controlled text generation~\cite{chen-etal-2019-controllable} tasks, such as paraphrasing~\cite{10.1162/tacl_a_00318,goyal-durrett-2020-neural} or controlled translation~\cite{DBLP:journals/corr/abs-2010-05856}. In this context, previous work~\cite{rashkin-etal-2019-towards} has also explored retrieval-based response selection. However, we are the first to explore joint retrieval and generative models, with the former guiding the latter.

\section{Method}
\label{sec:method}

The existing methods~\cite{li-etal-2020-empdg,lin-etal-2019-moel,majumder-etal-2020-mime} are strongly reliant on context emotion to generate empathetic responses. However, such approaches overlook the subtle expressions of understanding and personal resonance with the situation of the other interlocutor. To alleviate this, we take two major steps:
\begin{enumerate}
    \item We use exemplars to cue the generation on stylistic properties that signal empathy to the interlocutor
    \item We imbue the responses with four properties, namely sentiment, emotional presence, interpretation, and exploration, that are indicative of empathy~\cite{sharma2021facilitating}.
\end{enumerate}

\begin{figure*}
    \centering
    \includegraphics[width=0.85\linewidth]{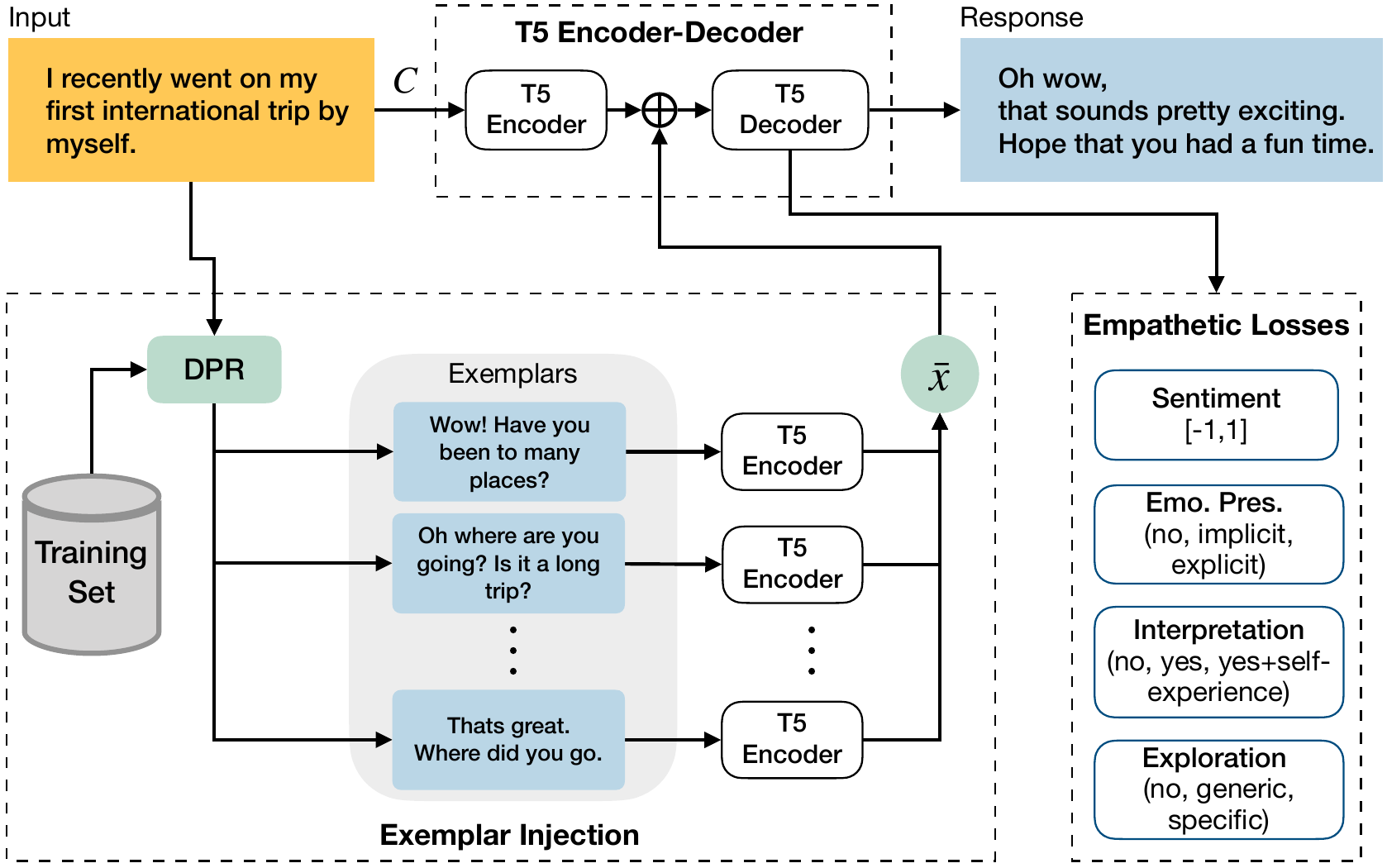}
    \caption{\footnotesize 
    The structure
    of our model \modelname.}
    \label{fig:model}
\end{figure*}

The exemplars for the former step are obtained by dense passage retrieval (DPR) \cite{karpukhin-etal-2020-dense} that extracts relevant exemplary responses from the training set for a given context. For the latter step, we firstly train individual classifiers that identify those four properties in a response. During the generative training the generated response is fed to these four classifiers. The four classification losses, based on synthetic labels, are back-propagated through these classifiers to train the generative encoder-decoder model. The resultant model, named \modelname, is visualized in \cref{fig:model}.

\subsection{Task Definition}

We are provided with a dialog between a user and an dialog agent, where the agents's role is to provide empathetic responses to the user's statements. Each turn in the dialog alternates between the user and the agent, which forms a sequence of utterances. For $n$ such utterances $[u_1, u_2, \dots, u_n]$ with the user emoting an emotion $e$, the goal of the agent (our model) is to empathetically respond to the last utterance $u_n$ from the user. In the mentioned sequence, all the odd-indexed $\{u_1, u_3, \dots, u_n\}$ and even-indexed utterances $\{u_2, u_4, \dots, u_{n-1}\}$ are uttered by the user and the empathetic agent, respectively. Thus for any instance, $n$ is always odd, i.e., the last turn in the provided dialog context is always that of the user. The goal of the task is to generate an appropriate and empathetic response---as the ${n+1}^{th}$ turn $u_{n+1}$---to this provided context.

\subsection{Encoder-Decoder Transformer}

The recent success of transfer learning from large self-supervised pre-trained models to myriad of NLP tasks had spurred us on to use text-to-text transfer transformer (T5)~\cite{JMLR:v21:20-074} as the basis of the encoder-decoder setup of our generative model. However, fine-tuning this pre-trained T5 model on this task counter-intuitively led to poor empathy in the responses. Hence, we resort to train the T5 model from scratch with GloVe~\cite{pennington-etal-2014-glove} word embeddings. We posit that this may be a side-effect of T5 having been pre-trained with plethora of text from the internet that are not particularly empathetic. Therefore, nudging such network to generate empathetic text may require larger number of samples. In \cref{sec:results_disc}, we empirically further explore this.

To this end, 
we first prepare the context by concatenating the contextual utterances $C=[u_1\oplus u_2\oplus \dots \oplus u_n]=[w_{11}, w_{12},\dots, w_{nm}]$, where $w_{ij}$ is the $j^\text{th}$ word in the $i^\text{th}$ utterance. The context $C$ is represented as the sum of word ($E_{W})$ and speaker embeddings~($E_S$): $E(C)=E_{W}(C)+E_S(C)$, where $E(C)\in \mathbb{R}^{k\times n_{emb}}$, $k$ is the number of words in $C$, and $n_{emb}$ is the embedding length of the words and speakers. The speaker embeddings are trained to represent the speakers of the utterances.

Context $E(C)$ is passed through the T5 encoder ($\TF_{enc}^{ctx}$) and subsequently through T5 decoder ($\TF_{dec}$) for empathetic response generation:
\begin{flalign}
Z&=\TF_{enc}^{ctx}(E(C)), \label{eqn:ctx_enc}\\
\mathcal{P}_{resp}&=\TF_{dec}(E_{W}(R_{1:t-1}), Z),
\end{flalign}
where $Z\in \mathbb{R}^{k\times D_{emb}}$ is the encoder output holding context representation, $R$ is the generated response of length $t$ ($R_0$ is \texttt{<start>} token), $\mathcal{P}_{resp}\in [0,1]^{t\times |V|}$ holds the probability distribution on each response token, and $V$ is the vocabulary.

In this work, we add add additional modules to this base encoder-decoder T5 transformer network to accommodate retrieved exemplars (\Cref{sec:exemplar_injection}) and four auxiliary losses based on \textit{sentiment}, \textit{emotional presence}, \textit{interpretation}, and \textit{exploration} (\Cref{sec:aux_losses}). 

\subsection{Exemplar Extraction and Injection}
\label{sec:exemplar_injection}

For each input context $C$, we fetch a list of possible responses from the training samples that are likely to embody context-specific empathetic characteristics. The key idea here is to reduce the reliance on emotions for empathetic response generation. Instead, we retrieve and show the model possible set of responses, called exemplars, for guidance. Needless to say, these responses are not fully aligned to the context, but they provided important cues towards generating the final response. For example, the exemplars can provide a filtered set of candidate responses with the appropriate empathetic phrases like \textit{I am sorry to hear that} if the user is sad about an event. We retrieve these exemplars using the popular DPR algorithm that we discuss next.

\paragraph{\textbf{Dense Passage Retrieval}}

Dense Passage Retrieval (DPR) is a recent algorithm proposed 
for
retrieval in a low-dimensional and continuous space~\cite{karpukhin-etal-2020-dense}. 
Unlike
traditional retrieval mechanisms, 
it
allows to effectively retrieve samples by matching the embedded representations over a large space of candidates. DPR is a bi-encoder that encodes text passages into a continuous embedding using an encoder such as BERT$_{BASE}$~\cite{DBLP:conf/naacl/DevlinCLT19}. Given a query, $C$ in our case, and a set of candidate responses from the training set $\Psi$'s, the goal of DPR is to learn a probability distribution $p(\Psi|C)$ according the some notion of similarity. This distribution is calculated as follows:
\begin{align}
    p(\Psi|C) = \exp({E_{cand}(\Psi)^TE_{query}(C)}),
\end{align}
where $E_{query}()$ and $E_{cand}()$ are the respective encoders of the context and candidate exemplars. Given this distribution, we can now choose the top-$q$ exemplars based on $p(\Psi|C)$. For this, DPR uses the Maximum Inner Product Search (MIPS) method---a sub-linear time approximated algorithm for nearest-neighbour search~\cite{DBLP:journals/corr/JohnsonDJ17}. We experiment with both the pre-trained DPR that is trained to retrieve answers for questions and also fine-tune DPR to our task. \Cref{sec:training_dpr} details the training process of the DPR.

\paragraph{\textbf{Injecting Exemplars}} We choose top-$q$ responses from the DPR as the exemplars for context $C$. To capture the necessary stylistic attributes, akin to the input context, each extracted exemplar ($\Psi_i$) is encoded with a T5 encoder ($\TF_{enc}^{exl}$) with identical architecture and GloVe word embedding:
\begin{flalign}
\mathcal{Z}_i = \TF_{enc}^{exl}(E_W(\Psi_i)).
\end{flalign}
The token-level exemplar representations $\mathcal{Z}_i\in \mathbb{R}^{\cdot\times n_{emb}}$ are mean-pooled to obtain a vector representation: $\psi_i = \text{mean}(\mathcal{Z}_i) \in \mathbb{R}^{n_{emb}}$. These representations from all the exemplars are aggregated to obtain the final exemplar representation:
\begin{flalign}
\chi=\text{mean}([\psi_1, \psi_2,\dots,\psi_q]).
\end{flalign}

To fuse information from the input context and the relevant stylistic cues from the extracted exemplars, exemplar representation $\chi$ is concatenated to the input context representation $Z$, from \cref{eqn:ctx_enc}, at token level and fed to a fully-connected layer $\text{FC}_{exl}$ of size $n_{emb}$:
\begin{flalign}
Z_{fused} = \text{FC}_{exl}([Z_i\oplus\chi]_{i=1}^{k}).
\end{flalign}
Subsequently, this fused representation $Z_{fused}\in \mathbb{R}^{k\times n_{emb}}$ is fed to the decoder for the response generation:
\begin{flalign}
&\mathcal{P}_{resp}=\TF_{dec}(E_{W}(R_{1:t-1}), Z_{fused}), \label{eqn:decode}\\
&p(R_i|C, R_{0:i-1}) = \mathcal{P}_{resp}[i].
\end{flalign}
The generative cross-entropy loss is defined as:
\begin{flalign}
L_{gen} = - \log p(R_{gold}|C),
\end{flalign}
where $R_{gold}$ is the reference response.

\subsection{Empathetic Losses on the Response}
\label{sec:aux_losses}

According to \citet{sharma2021facilitating}, emotional presence, interpretation, and exploration are important attributes of empathetic responses:
\begin{itemize}
\item
{\textbf{Emotional Presence.}} Is emotion in the response \emph{absent}, \emph{implicitly present}, or \emph{explicit present}?

\item
{\textbf{Interpretation.}} Does the response \emph{correctly} or \emph{incorrectly} interpret the user? In addition, does the response contain self-experiences?

\item
{\textbf{Exploration.}} Is there an attempt in the response to explore the user's emotion? \emph{no}, \emph{generically}, or \emph{specifically}.
\end{itemize}

To imbue the generated response with the aforementioned qualities, corresponding synthetic labels are obtained from the gold responses. To this end, we train classifiers --- identical in structure to \cref{eqn:aux_out}, but with independent parameters --- for each of these three attributes with the \texttt{EmpathyMentalHealth} dataset\footnote{https://bit.ly/2Rwy2gx}. In the \texttt{EmpathyMentalHealth} dataset, each of these attributes is further categorized on a scale of 1/2/3 indicating low/mid/high levels. Hence, the classifier for each of these three attributes is trained with a three class classification objective. We provide more details about this training procedure in \cref{sec:synthetic}. The gold responses from \texttt{EmpatheticDialogues} are fed to these trained classifiers to obtain the synthetic labels. 

Following the prior works~\cite{majumder-etal-2020-mime,lin-etal-2019-moel}, we additionally assume sentiment an important factor in empathy:
\begin{itemize}
\item
{\textbf{Sentiment.}} VADER~\cite{vader} is used to generate a sentiment score in $[-1, 1]$ for each response; 
$-1$, $1$, and $0$ being extreme negative, extreme positive, and neutral, respectively.
\end{itemize}

Final hidden state of decoder $\TF_{dec}$ (\cref{eqn:decode}) is taken as the generated-response representation
$\Tilde{R}\in \mathbb{R}^{t\times n_{emb}}$. For each of the four attributes (denoted as *), 
we consider the starting token vector $\Tilde{R}[0]$ and fed to a classifier:
\begin{flalign}
\mathcal{P}_{*} &= \text{softmax}(\text{FC}_*(\Tilde{R}[0])).
\label{eqn:aux_out}
\end{flalign}
For \emph{sentiment}, tanh activation is used instead of softmax in \cref{eqn:aux_out}.

We use categorical cross-entropy as loss for \emph{emotional presence}, \emph{interpretation}, and \emph{exploration}: $L_{EP}$, $L_{int}$, and $L_{exp}$, respectively. Mean squared error is used as loss for \emph{sentiment}: $L_{sent}$. 
\cref{algo:empsen} depicts the algorithm used to pre-train the classifiers for empathetic losses and how we use them in our end-to-end generative framework i.e., \modelname{}.

\subsection{Training}

We use Adam~\cite{DBLP:journals/corr/KingmaB14} to optimize the overall loss:
\begin{flalign*}
\mathcal{L} = \alpha_{gen}L_{gen} + \alpha_{EP} L_{EP}  + \alpha_{int} L_{int}  + \alpha_{exp} L_{exp} + \alpha_{sent} L_{sent},
\end{flalign*}
where $\alpha_*$ are hyper-parameters to assign importance to constituent losses. Please check \cref{algo:lempex} and \cref{algo:lempexinference} for a brief overview of the algorithms used for training \modelname{} and doing inference using it. All the models were trained for 50 epochs; early-stopping was used with the patience of 5; the learning rate was set to 1e-5. We used a batch size of 8 for the pre-trained models and a batch size of 32 was used for the Glove-based models. $\alpha_{gen}$ was set to 1.0; $\alpha_{EP}, \alpha_{int}, \alpha_{exp}$, and $\alpha_{sent}$ were assigned to 0.1. The lower importance of the empathetic losses is noteworthy; assignment of higher importance resulted in empathetic, yet, vapid and repetitive responses in our experiments. The hyperparameters were fixed using a random search on the validation dataset.

\begin{algorithm}[ht!]
\SetAlgoLined
 Training Dataset $\mathcal{D}$ = \texttt{EmpathyMentalHealth} Dataset\;
 Initialize $T5_{Enc}$ as empathy model \textbf{E}\; 
 \textbf{E} is the compact representation of three different models for three empathy dimensions -- Emotional Presence, Interpretation, and Exploration\;
 Initialize $T5_{Enc}$ as sentiment model \textbf{S}\;
 \For{dialogue id \boldmath{$d$} with context \boldmath{$c$}, response \boldmath{$r$} and empathy label \boldmath{$y_e$} in $\mathcal{D}$}{
  Gold sentiment of \boldmath{$r$}: \boldmath{$y_s$} = \textbf{VADER}(\boldmath{$r$}) \;
  
  Encode and predict empathy: \\
  $\qquad$ \boldmath{$\hat{y}_e$} = \textbf{$C$}(\boldmath{$[c, r]$})\;
  
  Encode and predict sentiment: \\
  $\qquad$ \boldmath{$\hat{y}_s$} = \textbf{$S$}(\boldmath{$r$})\;
  
  Compute loss: \\
  $\qquad$ \boldmath{$l_e$} = Loss(\boldmath{$y_e$}, \boldmath{$\hat{y}_e$})\;
  $\qquad$ \boldmath{$l_s$} = Loss(\boldmath{$y_s$}, \boldmath{$\hat{y}_s$})\;
  
  Backpropagate loss \boldmath{$l_e$} to update \textbf{E} \;
  Backpropagate loss \boldmath{$l_s$} to update \textbf{S} \;
 }
 \caption{Empathy and Sentiment model Training Algorithm}
 \label{algo:empsen}
\end{algorithm}

\begin{algorithm}[ht!]
\SetAlgoLined
 Training Dataset $\mathcal{D}$ = \texttt{EmpatheticDialogues} Training Split\;
 Initialize pre-trained BERT-Base as context encoder \textbf{C}\;
 Initialize pre-trained BERT-Base as response encoder \textbf{R}\;
 Dot product similarity function \boldmath{$sim(x, y) = x.y$} \;
 \For{dialogue id \boldmath{$d$} with context \boldmath{$c$}, response \boldmath{$r$} and emotion \boldmath{$e$} in $\mathcal{D}$}{
  Assign positive response: \\
  $\qquad$ \boldmath{$r^+$} = \boldmath{$r$}
  
  Assign negative responses: \\
  $\qquad$ $\mathcal{D}_1 = \mathcal{D} \setminus d $ with emotion=\boldmath{$e$} \;
  $\qquad$ $\mathcal{D}_2 = \mathcal{D} \setminus d $ with emotion$\neq$\boldmath{$e$} \;
  $\qquad$ \boldmath{$r^-_{1}$} = random sample of responses from $\mathcal{D}_1$\;
  $\qquad$ \boldmath{$r^-_{2}$} = random sample of responses from $\mathcal{D}_2$\;
  $\qquad$ \boldmath{$r^-$} = \boldmath{$r^-_{1} \bigcup r^-_{2}$}
  
  Compute exponentiated similarities: \\
  $\qquad$ \boldmath{$s^+$} = $exp(sim(\textbf{C}(\boldmath{c}), \textbf{R}(\boldmath{r^+})))$ \;
  $\qquad$ \boldmath{$s^-$} = $\sum_{j=1}^{n}exp(sim(\textbf{C}(\boldmath{c}), \textbf{R}(\boldmath{r^-_{j}})))$ \;
  
  Compute negative log-likelihood of positive response: \\
  $\qquad$ \boldmath{$l$} = \boldmath{$-log (\frac{s^+}{s^+ + s^-}) $} \;
  
  Backpropagate loss \boldmath{$l$} to update \textbf{C}, \textbf{R}\;
 }
 \caption{DPR Training Algorithm}
 \label{algo:dprtraining}
\end{algorithm}

\begin{algorithm}[ht!]
\SetAlgoLined
 Dataset $\mathcal{D}$ = \texttt{EmpatheticDialogues} Training Split\;
 Collect all responses $\mathcal{R}$ from $\mathcal{D}$: \\
  $\qquad \mathcal{R} = \{r_1, r_2, .., r_n\}$ \;
  
 \For{dialogue id \boldmath{$d$} with context \boldmath{$c$} in train/val/test dataset}{
  
  Compute similarities with $\mathcal{R}$: \\
  \For{j=1 to n}{
  \boldmath{$s_j$} = $sim(\textbf{C}(\boldmath{c}), \textbf{R}(\boldmath{r_j}))$ \;
  }
  \boldmath{$s=[s_1, s_2, .., s_j]$} \;
  Sort \boldmath{s} in decreasing order \; 
  Return responses corresponding to first \boldmath{k} elements of sorted \boldmath{s} \;
 }
 \caption{DPR Inference Algorithm}
 \label{algo:dprinference}
\end{algorithm}

\subsubsection{\textbf{Training the DPR}}
\label{sec:training_dpr}

The pre-trained DPR model is fine-tuned on a dataset based on the training set of \texttt{EmpatheticDialogues}.
Each training sample for DPR consists of an input context and a set of responses that contains exactly one positive response, which is the gold response, and $n_{neg}$ number of negative responses $R^-_i$,
which are randomly sampled from the utterances of the other 
training dialogues with different emotion: 
\begin{flalign*}
\overbrace{[u_1, u_2,\dots,u_j]}^{\text{Context}}, [\underbrace{R_{gold}}_{\text{Positive Response}}, \overbrace{R^-_1, R^-_2,\dots, R^-_{n_{neg}}}^{\text{Negative Responses}}]
\end{flalign*}
DPR is fined-tuned to predict the correct positive sample from the supplied set of responses. It first encodes the context and responses into a continuous space using two separate encoders termed as bi-encoders. The bi-encoder training is then performed by simultaneously maximizing the similarity between the context and the gold response, and minimizing the similarity between the context and the negative responses. Dot product is used as a measure of similarity and the negative log-likelihood of the positive response is optimized as the objective function:
\begin{flalign}
&L_{dpr} = - \log \frac{e^{sim(c, r_{gold})}}{e^{sim(c, r_{gold})} + \sum_{j=1}^{n_{neg}} sim(c, r^-_j)},
\end{flalign}
where $sim(c, r) = c^Tr$ and $c, r_{gold}, r^-_j$ are vectors corresponding to context, positive response, and negative responses, respectively, encoded by the DPR bi-encoders. The similarity function $sim(c, r)$ computes the dot-product between the context vector $c$ and the (positive or negative) response vector $r$. 

We fine-tune the DPR model with our examples starting from the \textit{Natural Questions BERT Base} checkpoint \cite{karpukhin-etal-2020-dense}. This model uses BERT$_{BASE}$ as bi-encoders and has been pretrained on the Natural Questions dataset \cite{kwiatkowski2019natural} for open-domain question answering.

During inference for exemplar extraction, however, for each input context a set of responses from the training dialogues of \texttt{EmpatheticDialogues} with the same emotion is fed to DPR. The top $q$ responses with the highest DPR-based dot product similarity with the context are chosen as exemplars. Notably, the set of exemplar responses does not contain the gold response or any other response from the same dialogue for the training samples. In other words, exemplar responses are always selected from a different dialogue (but belonging to the same emotion category). Please check \cref{algo:dprtraining} and \cref{algo:dprinference} for a brief overview of the algorithms used for DPR training and inference.

\begin{algorithm}[ht]
\SetAlgoLined
 Train empathy and sentiment prediction models: \textbf{E} and \textbf{S} using \cref{algo:empsen}. \textbf{E} is the compact representation of three different models as in \cref{algo:empsen} \;
 Train \textbf{DPR} with positive and negative responses using \cref{algo:dprtraining}\;
 \For{context \boldmath{$c$} and response \boldmath{$r$} in training set}{
  Create synthetic gold labels: \\
  $\qquad$ \boldmath{$g_e, g_s$} = \textbf{E}(\boldmath{\boldmath{$c$}, $r$}), \textbf{S}(\boldmath{$r$})\;
  
  Retrieve exemplars: \\
  $\qquad$ \boldmath{$e_1, .., e_n$} = \textbf{DPR}(\boldmath{$c$})\;
  
  Compute exemplar representation: \\
  $\qquad$ \boldmath{$\hat{e}$} = $\frac{1}{n}\sum_{i=1}^{n}$\textbf{$T5_{Enc}$}(\boldmath{$e_i$})\;
    
  Encode context: \\
  $\qquad$ \boldmath{$\hat{c}$} = \textbf{$T5_{Enc}$}(\boldmath{$c$})\;
  
  Transform representation: \\
  $\qquad$ \boldmath{$\hat{x}$} = \textbf{Linear}([\boldmath{$\hat{e}$}, \boldmath{$\hat{c}$}])\;
  
  Output response token probabilities: \\
  $\qquad$ \boldmath{$\hat{r}$} = \textbf{$T5_{Dec}$}(\boldmath{$\hat{x}$})\;
  
  Obtain empathy, sentiment predictions: \\
  $\qquad$ \boldmath{$\hat{g}_e, \hat{g}_s$} = \textbf{E}(\boldmath{$c$}, \boldmath{$\hat{r}$}), \textbf{S}(\boldmath{$\hat{r}$})\;
  
  Compute loss: \\
  $\qquad$ \boldmath{$l_1$} = Loss(\boldmath{$r$}, \boldmath{$\hat{r}$})\;
  $\qquad$ \boldmath{$l_2$} = Loss(\boldmath{$g_e$}, \boldmath{$\hat{g_e}$})\;
  $\qquad$ \boldmath{$l_3$} = Loss(\boldmath{$g_s$}, \boldmath{$\hat{g_s}$})\;
  $\qquad$ \boldmath{$l$} = \boldmath{$\alpha*l_1+\beta*l_2+\gamma*l_3$}\;
  
  Backpropagate loss \boldmath{$l$} to update \textbf{$T5_{Enc}$}, \textbf{$T5_{Dec}$} \;
 }
 \caption{LEMPEx Training Algorithm}
 \label{algo:lempex}
\end{algorithm}

\begin{algorithm}[ht!]
\SetAlgoLined
 \For{context \boldmath{$c$} in test set}{
 
  Retrieve exemplars (using \cref{algo:dprinference}) from the training fold of \texttt{EmpatheticDialogues} dataset using the trained DPR model (see \cref{algo:dprtraining}): \\
  $\qquad$ \boldmath{$e_1, .., e_n$} = \textbf{DPR}(\boldmath{$c$})\;
  
  Compute exemplar representation: \\
  $\qquad$ \boldmath{$\hat{e}$} = $\frac{1}{n}\sum_{i=1}^{n}$\textbf{$T5_{Enc}$}(\boldmath{$e_i$})\;
    
  Encode context: \\
  $\qquad$ \boldmath{$\hat{c}$} = \textbf{$T5_{Enc}$}(\boldmath{$c$})\;
  
  Transform representation: \\
  $\qquad$ \boldmath{$\hat{x}$} = \textbf{Linear}([\boldmath{$\hat{e}$}, \boldmath{$\hat{c}$}])\;
  
  Generate response with beam-search: \\
  $\qquad$ \boldmath{$\hat{y}$} = \textbf{$Beam Search (T5_{Dec}$}(\boldmath{$\hat{x}$}))\;
  
 }
 \caption{LEMPEx Inference Algorithm}
 \label{algo:lempexinference}
\end{algorithm}

\section{Experimental Settings}

\subsection{Dataset} \label{sec:empathetic_dataset}

We evaluate our method on \texttt{EmpatheticDialogues}\footnote{https://github.com/facebookresearch/EmpatheticDialogues}~\cite{rashkin-etal-2019-towards} dataset. It contains $24,850$ dyadic dialogues where the interlocutors empathetically engage with each other. The samples consist of a context, that is an excerpt from a conversation, and a response to the last utterance of the context. All the user utterances of a dialogue are annotated with an emotion label out of total 32 emotion labels. The emotion labels are uniformly distributed across the dataset. The dialogues are split by 8:1:1 (training:validation:test) ratio that is defined by the curators of the dataset.

\subsection{Baseline Methods and State of the Art}

\paragraph{\textbf{MIME}~\cite{majumder-etal-2020-mime}} This method aims at systematically mimicking the overall sentiment of the user in the response to generate empathy. To this end, it splits the set of emotions into two groups having positive and negative emotions. The final response is decoded based on an attention-based blend of these two emotion groups.

\paragraph{\textbf{EmpDG}~\cite{li-etal-2020-empdg}} EmpDG separately considers the semantic and emotional context during encoding. For response decoding, on the other hand, semantic and emotional representations are fused. It further employs semantic and emotion discriminators that minimize the Wasserstein-1 distance between the semantic and emotional aspects of gold and generated response. This is supposed to instill further semantic and emotional perception into the response. This method is considered as the state of the art (SOTA).

\paragraph{\textbf{T5}~\cite{JMLR:v21:20-074}} This baseline has two variants:
\begin{itemize}
    \item \emph{\textbf{T5-PT}} is a fine-tuned version of a checkpoint of T5-small model trained on C4 dataset~\cite{JMLR:v21:20-074};
    \item \emph{\textbf{T5-GloVe (\modelname{} w/o exemplars, empathetic losses)}} is a T5-small model, with GloVe~\cite{pennington-etal-2014-glove} word embeddings, trained from scratch on \texttt{EmpatheticDialogues} dataset. 
    It
    is considered for 
    fair comparison with the existing GloVe-based methods, such as MIME and EmpDG. Notably, 
    it
    is 
    equivalent to an ablated \modelname{} without exemplars and empathetic~losses.
\end{itemize}

\subsection{Evaluation}

We employ both automatic and human-based metrics to evaluate and compare our method and the baselines:

\subsubsection{\textbf{Automatic Metrics}}
Similar to previous works~\cite{li-etal-2020-empdg,majumder-etal-2020-mime,lin-etal-2019-moel}, we consider the following automatic metrics:

\begin{itemize}[wide, labelwidth=!, labelindent=0pt]
    \item \emph{\textbf{BLEU}} score~\cite{Papineni02} has long been used to compare generated text against references in language generation tasks. However, it has been shown to bear little correlation with human evaluation in many dialogue generation tasks, including empathetic response generation~\cite{lin-etal-2019-moel,majumder-etal-2020-mime}. As such, we just keep it as a reference rather than using as a mode of comparison.
    \item \emph{\textbf{Perplexity}}, defined as $e$ raised to the power of cross-entropy, is also kept as reference.
    \item \emph{\textbf{Diversity}} is meant to quantify lexical variety in the responses~\cite{DBLP:conf/naacl/LiGBGD16}. Specifically, \emph{Distinct-1} and \emph{Distinct-2} metrics reflect the variation of unigrams and bigrams, respectively, in the responses:
    \begin{flalign*}
    \text{Distinct-}n = \frac{\#(\text{unique }n\text{-grams})}{\#(\text{tokens})} 100 \%.
    \end{flalign*}
\end{itemize}

\subsubsection{\textbf{Automatic Synthetic Label-Based Evaluation}}
\label{sec:synthetic}
~\\
We use the \texttt{EmpathyMentalHealth}~\cite{sharma2020empathy} dataset to create synthetic labels for the responses in \texttt{Empathetic Dialogues} dataset. The \texttt{EmpathyMentalHealth} contains (context, response) pairs with annotated \emph{emotional presence, interpretation, exploration} labels. Labels for all the dimensions are provided on a scale of 1/2/3 indicating low/mid/high levels. 

We consider three different classification models for \emph{emotional presence, interpretation}, and \emph{exploration} dimensions.
We use the pretrained T5 model \cite{raffel2020exploring} having only the encoder part as the backbone of the classification models. A linear layer is added on top of the encoders with softmax activation for the three class classification.

The classification models are trained on the (context, response, label) triplets. We concatenate the context and response pair and pass it as input to the T5 encoder model. 
The label is then classified from the final layer vector corresponding to the starting token \textit{<s>}.

We use 20\% of the annotations in \texttt{EmpathyMentalHealth} dataset as the validation split. Results obtained in this validation set for the classification of \emph{emotional presence, interpretation}, and \emph{exploration}
are shown in \cref{tab:emh}. Results indicate that the \emph{exploration} dimension can be predicted most accurately with close to 94\% weighted-F1 (W-F1) score. The \emph{emotional presence} and \emph{interpretation} labels are more difficult to predict, with scores being close to 83\% and 84\%, respectively.

We consider the checkpoints obtaining the best results in the validation set, and use them to create the synthetic labels for the datapoints in \texttt{EmpatheticDialogues}. For a given instance, we concatenate the context $C$ and reference response $R_{gold}$, and pass it to the trained models and use the predicted class as the gold synthetic label. Additionally, we create gold sentiment scores of the reference responses using the VADER tool.

For a generated response, we follow the same method to predict the \emph{emotional presence, interpretation, exploration} classes and the 
\emph{sentiment} score. We then perform automatic evaluations based on the gold synthetic labels/scores and predicted labels/scores. For classification,  we consider the macro-F1 between gold and predicted labels. For regression, we consider the mean absolute error (mae) between gold and predicted scores.

\subsubsection{\textbf{Human Ratings}}
We randomly choose 100 samples from the testing set and employ three human annotators to evaluate the corresponding responses by different methods.
We utilize two sets of attributes to be evaluated by human annotators, namely, \emph{coarse attributes} and finer \emph{empathy-based attributes}.


The human annotators rate on a integer scale from 1 (worst) to 5 (best) on the three \emph{coarse attributes}:
\begin{itemize}[wide, labelwidth=!, labelindent=0pt]

    \item 
    \textbf{Empathy:} How empathetic is the response to the user?
    
    \item 
    \textbf{Relevance:} How relevant is the response to the context and user?
    
    \item 
    \textbf{Fluency:} How linguistically intelligible is the response?
\end{itemize}
\begin{table}[t!]
\small
\centering
\caption{\footnotesize 
Accuracy (Acc) and weighted-F1 (W-F1) scores of 3-way classification for emotional presence, interpretation, and exploration prediction in validation set of \texttt{EmpathyMentalHealth}.}
\label{tab:emh}
\scalebox{0.95}
{
\begin{tabular}{lrrrrrr}
\toprule
Dimension & Acc & W-F1 \\
\midrule
Emotional Presence & 82.99 & 82.89 \\
Interpretation & 84.47 & 83.70 \\
Exploration & 94.04 & 93.92 \\
\bottomrule
\end{tabular}
}
\end{table}
As per \citet{sharma2020empathy}, the three \emph{empathy-based attributes} are strongly related to empathy (elaborated in \cref{sec:intro}):

\begin{itemize}[wide, labelwidth=!, labelindent=0pt]
    \item \textbf{Emotional Presence:} 
        \begin{enumerate}[label={\arabic* -}, start=1]
            \item The response has no emotion, or has emotions that are contradictory to user's feeling, or apathy.
            \item The response \textit{implicitly} alludes to certain feelings of the user.
            \item The response \textit{explicitly} alludes to certain feelings of the user.
        \end{enumerate}
        
    \item \textbf{Interpretation:} 
        \begin{enumerate}[label={\arabic* -}, start=1]
            \item The response communicates no understanding or misunderstanding of the user.
            \item The response communicates correct understanding of the user's situation.
            \item The response also mentions of self-experiences of the agent to amplify the agent's understanding of user's situation.
        \end{enumerate}
        
    \item \textbf{Exploration:} 
        \begin{enumerate}[label={\arabic* -}, start=1]
            \item The response makes no attempt to explore the user's situation or the exploration attempt is invalid in the context.
            \item Exploration attempt is generic.
            \item Exploration attempt is specific to the context.
        \end{enumerate}
\end{itemize}

These attribute scores across all 100 samples and three annotators are averaged to obtain the overall evaluation of a particular model.
\begin{table*}[t]
\small
    \centering
    \caption{\footnotesize Comparing efficacy of our model \modelname{} against the baseline models on various automated and human-evaluated metrics.}
    \resizebox{0.8\linewidth}{!}{
    \begin{tabular}{|l|l|cc|ccc|ccc|cc|}
    \hline
     \multirow{2}{*}{\textbf{Method}} & \multirow{2}{*}{\textbf{\#params.}} & \multirow{2}{*}{\textbf{BLEU}} & \multirow{2}{*}{\textbf{PPL}} & \multicolumn{3}{c|}{\it \textbf{Coarse Attributes}} & \multicolumn{3}{c|}{\it \textbf{Empathy-Based Attributes}} & \multicolumn{2}{c|}{\it \textbf{Diversity} (\%)} \\
     &&&& Empathy & Relevance & Fluency & \makecell[c]{Emotion\\Presence} & \makecell[c]{Interpretation} & Exploration & Distinct-1 & Distinct-2 \\
        \hline
        Gold && 100 & -- & 4.48 & 4.98 & 4.98 & 1.95 & 1.92 & 2.34 & 7.94 & 42.70 \\
       \hline
       MIME & 16.95M & 8.76 & 37.33 & 3.00 & 3.15 & 4.39 & 1.75 & 1.4 & 1.22 & 0.63 & 3.97 \\
       EmpDG (SOTA) & 28.38M  & 8.61 & 34.18 & 2.95 & 3.1 & 4.32 & 1.64 & 1.3 & 1.69 & \bf 1.81 & 6.38 \\
       T5-PT & 60.50M &  7.47 & 36.88 & 3.31 & 3.75 & \bf 4.78 & 1.61 & 1.49 & 1.82 &  1.59 & \bf 17.33 \\
       \hline
        \makecell[l]{T5-GloVe/\modelname{} \\
            \quad w/o Exemplars, \\ \quad w/o Emp. Losses} & 27.57M & 7.38 & \bf 25.26 & 3.34 & 3.36 & 4.42 & 1.84 & 1.43 & 1.68 & 1.37 & 13.72\\
       \begin{tabular}{@{}l}
            \modelname{}  \\
            \quad w/o Emp. Losses  
         \end{tabular} & 165.33M & 7.49 & 26.88 & 3.55 & 3.69 & 4.65 & 1.88 & 1.52 & 1.76 & 1.39 & 14.55 \\
       \modelname{} & 165.33M & \bf 7.88 & 26.37 & \bf 3.76 & \bf 3.78 & 4.61 & \bf 2.02 & \bf 1.64 & \bf 2.02 & 1.41 & 14.66\\
        \hline
    \end{tabular}
    }
    \label{tab:gen_results}
\end{table*}
\subsubsection{\textbf{A/B Testing}}

In addition to the Likert-based ratings described above, we also perform ranking-based human evaluation. Given a dialog context and two candidate responses from Model A and Model B, we ask three human annotators to chose their preferred response between the two choices. They may choose to vote \texttt{tie} if both responses are deemed equal. In our experiment, this evaluation was performed on the same set of 100 test samples as for human ratings. We apply \textit{majority voting} to aggregate votes from the three annotators. In case of three distinct votes from three annotators, a fourth annotator is brought in to break the tie. Finally, we calculate the percentage of times model A is preferred (WIN), model B is preferred (LOSS), and A and B tie (TIE).

\section{Results and Discussions}
\label{sec:results_disc}

\subsection{Generation Performance}

\begin{table*}[ht!]

\centering
\caption{
\footnotesize Top exemplars from the DPR model fine-tuned on Empathetic Dialogs, and the original pre-trained DPR checkpoint without any further training. The exemplars from the fine-tuned DPR model are considerably more empathetic, diverse and contextually relevant.
Notably, exemplars from the fine-tuned DPR are not always semantically similar to the references, although they are stylistically plausible and relevant with respect to the context.}
\begingroup
\setlength{\tabcolsep}{6.5pt} 
\renewcommand{\arraystretch}{1.1} 
\resizebox{\linewidth}{!}
{\begin{tabular}{|L{8.5cm}|L{7cm}|L{4cm}|L{4cm}|}
\hline
\textbf{Context} & \textbf{Reference} & \textbf{Exemplar Fine-Tuned DPR} & \textbf{Exemplar Pre-Trained DPR}\\
\hline
I was going through some boxes the other day. I found some old pictures of my kids I thought were gone. & That is exciting! I love having pictures to look back on. & Pictures are nice to have to be able to look back on life. & Those are nice memories.\\
\hline
I recently hit a new personal record at the gym! I was very excited. & That is a great accomplishment. I admire your dedication in obtaining that goal. & Whoa, that is really impressive! How much do you bench now? & Ohh nice!\\
\hline
I live alone. Sometimes it can be hard with nobody around. & That must feel lonely.  You should go out and try to meet some people! & I can imagine that. I've never lived alone before. & How are you keeping yourself occupied? \\
\hline
I wish my neighbors were more considerate. They woke me up at 4am recently. & Wow did you call someone about that? & That was rude. Did you confront them? & Oh yeah? Why is that?  \\
\hline
We just found out the person working in our accounting department has been stealing money from our company for years. & Yikes! That is terrible! What a awful thing to do! I am glad he finally got busted! Wow, for years, huh? & Wow, that is pretty crazy! What did you do? & I would have screamed!\\
\hline
\end{tabular}
}
\endgroup
\label{tab:exemplars}
\end{table*}

As per \cref{tab:gen_results}, our model \modelname{} outperforms the state of the art (SOTA) on all human-evaluated metrics by a significant margin. Delving deeper, T5-GloVe, a variant of \modelname{} itself, beats the SOTA on all the human-evaluated metrics, save for \emph{exploration} dimension. Although, the difference is imperceptible there. This might be attributed to the six-layered transformer encoder and decoder in T5 as opposed to EmpDG and MIME that use single-layered transformer encoder and decoder. 
The strong baselines MIME and EmpDG perform significantly worse than \modelname{} according to both automatic and human evaluation. We also annotated 100 samples of the gold dataset to set up a fair comparative baseline. The results 
in \cref{tab:gen_results} 
show 
that all the models are far from matching the results of the gold standard data. As expected, the pre-trained T5 model---\emph{T5-PT} outperforms all other methods in terms of fluency. \modelname{} further encapsulates the empathy-specific losses and its performance for empathy score as per the human evaluation has surged by a significant margin. 

However, this improvement comes at a cost of a slight decrease in the fluency score when the empathy-specific losses are introduced to \modelname. \modelname{} also outshines other models in the empathy-based human evaluation. Results suggest that retrieved exemplars and empathy-specific losses have a notable impact on the performance of \modelname{} in this evaluation scheme. According to the human evaluators, \modelname{} excels in picking up the right emotional keyword for empathy, and it produces more relevant contextual words. Apart from attaining comparatively better BLEU and perplexity scores, \modelname{} also obtains significantly better Distinct-1 and Distinct-2 scores which indicate the superiority of \modelname{} to its competitors in generating diverse responses.
\begin{table*}[t!]
\small
\centering
\begingroup
\renewcommand{\arraystretch}{1.1} 
\caption{\footnotesize Comparing responses between models.
\vspace{-1ex}
}
\label{tab:case_study}
\resizebox{0.85\linewidth}{!}
{\begin{tabular}{|l|l|ccccc|}
\hline
& & \textbf{Relevance} & \makecell{\textbf{Emotional}\\\textbf{Presence}} & \textbf{Interpretation} & \textbf{Exploration} & \textbf{Winner} \\
\hline
\hline
\textbf{Emotion} & \multicolumn2l{\textsc{TERRIFIED}} & & & & \\
\textbf{Context} & \multicolumn5l{My dad called me one day crying out for help. I had to drive over to his house.} &   \\
\textbf{Reference} & \multicolumn2l{Oh my goodness! What happened?} & & & & \\
\cline{2-7}
\makecell[r]{\colorbox{brilliantlavender}{\textsc{\modelname}}} & Oh no, did he get hurt? I hope it wasn't too bad. & \cellcolor{Green2} High & \cellcolor{Green1} Mid & \cellcolor{Green2} High & \cellcolor{Green2} High & \multirow{4}{*}{\makecell[r]{\colorbox{brilliantlavender}{\textsc{\modelname}}}} \\
\makecell[r]{\colorbox{ghostwhite}{\textsc{T5-GloVe}}} & That's awful. Is he okay? Does it have someone you can do that & \cellcolor{Green1} Mid & \cellcolor{Green1} Mid & \cellcolor{Green1} Mid & \cellcolor{Green2} High & \\
\makecell[r]{\colorbox{Yellow1}{\textsc{MIME}}} & That is good to hear. How did he die? & \cellcolor{Green1} Mid & \cellcolor{Gray1} NIL & \cellcolor{Gray1} NIL & \cellcolor{Gray1} NIL & \\
\makecell[r]{\colorbox{Blue1}{\textsc{EmpDG}}} & That is so sweet. I am sure you are so hard. & \cellcolor{Gray1} NIL & \cellcolor{Gray1} NIL &  \cellcolor{Gray1} NIL & \cellcolor{Gray1} NIL & \\

\hline
\hline

\textbf{Emotion} & \multicolumn2l{\textsc{CONFIDENT}} & & & & \\
\textbf{Context} & \multicolumn5l{I know I shouldn't count my chicks before they hatch, but I'm feeling pretty good after that interview today!} & \\
\textbf{Reference} & \multicolumn2l{Oh what job was the interview for?} & & & & \\
\cline{2-7}
\makecell[r]{\colorbox{brilliantlavender}{\textsc{\modelname}}} & That is great. You must have felt proud of yourself! I am sure you will get it. & \cellcolor{Green1} Mid & \cellcolor{Green2} High & \cellcolor{Green2} High & \cellcolor{Gray1} NIL & \multirow{4}{*}{\makecell[r]{\colorbox{brilliantlavender}{\textsc{\modelname}}}}\\
\makecell[r]{\colorbox{ghostwhite}{\textsc{T5-GloVe}}} & Oh wow. I bet the zoo is really amazing! You can do it! & \cellcolor{Gray1} NIL & \cellcolor{Green1} Mid & \cellcolor{Gray1} NIL & \cellcolor{Gray1} NIL & \\
\makecell[r]{\colorbox{Yellow1}{\textsc{MIME}}} & I bet you are so proud of yourself! & \cellcolor{Green1} Mid & \cellcolor{Green1} Mid & \cellcolor{Green2} High & \cellcolor{Gray1} NIL & \\
\makecell[r]{\colorbox{Blue1}{\textsc{EmpDG}}} & What kind of I do you think you will do? & \cellcolor{Gray1} NIL & \cellcolor{Gray1} NIL & \cellcolor{Gray1} NIL & \cellcolor{Gray1} NIL & \\

\hline
\hline

\textbf{Emotion} & \multicolumn2l{\textsc{SENTIMENTAL}} & & & &  \\
\textbf{Context} & \multicolumn2l{I was heartbroken when my dad passed away.. still not over it.} & & & & \\
\textbf{Reference} & \multicolumn5l{Sorry to hear that... they say time heals everything but i disagree its been 9 years for me and it still feels like yesterday. How long for you?} & \\
\cline{2-7}
\makecell[r]{\colorbox{brilliantlavender}{\textsc{\modelname}}} & I'm sorry to hear that. It's so hard to lose a friend. & \cellcolor{Green2} High & \cellcolor{Green2} High & \cellcolor{Green2} High & \cellcolor{Gray1} NIL & \multirow{4}{*}{\makecell[r]{\colorbox{brilliantlavender}{\textsc{\modelname}}}}\\
\makecell[r]{\colorbox{ghostwhite}{\textsc{T5-GloVe}}} & Oh that's terrible . What happened? Your aunt's wrong ? & \cellcolor{Green1} Mid & \cellcolor{Green1} Mid & \cellcolor{Gray1} NIL & \cellcolor{Gray1} NIL & \\
\makecell[r]{\colorbox{Yellow1}{\textsc{MIME}}} & Oh, I am sorry. I hope he was a good thing to do. &\cellcolor{Green0} Low & \cellcolor{Green2} High & \cellcolor{Gray1} NIL & \cellcolor{Gray1} NIL & \\
\makecell[r]{\colorbox{Blue1}{\textsc{EmpDG}}} & I am so sorry to hear that. &\cellcolor{Green0} Low & \cellcolor{Green2} High & \cellcolor{Gray1} NIL & \cellcolor{Gray1} NIL & \\

\hline
\hline

\textbf{Emotion} & \multicolumn2l{\textsc{ANNOYED}} & & & & \\
\textbf{Context} & \multicolumn2l{I left my food on the coffee table and my dog ate it.} & & & &  \\
\textbf{Reference} & \multicolumn5l{That is a terrible tragedy, my friend. Did the dog get sick?} &  \\
\cline{2-7}
\makecell[r]{\colorbox{brilliantlavender}{\textsc{\modelname}}} & That sucks. You must have been angry. That happens to me. & \cellcolor{Green1} Mid & \cellcolor{Green2} High & \cellcolor{Green2} High & \cellcolor{Gray1} NIL & \multirow{4}{*}{\makecell[r]{\colorbox{brilliantlavender}{\textsc{\modelname}}}}\\
\makecell[r]{\colorbox{ghostwhite}{\textsc{T5-GloVe}}} & Oh that' s terrible, I hope you can find a friend! &\cellcolor{Green0} Low & \cellcolor{Green1} Mid & \cellcolor{Green1} Mid & \cellcolor{Gray1} NIL & \\
\makecell[r]{\colorbox{Yellow1}{\textsc{MIME}}} & That sounds really frustrating. What happened? & \cellcolor{Green1} Mid & \cellcolor{Green1} Mid & \cellcolor{Green1} Mid & \cellcolor{Gray1} NIL & \\
\makecell[r]{\colorbox{Blue1}{\textsc{EmpDG}}} & Oh no! What happened? & \cellcolor{Green1} Mid  & \cellcolor{Green1} Mid & \cellcolor{Gray1} NIL & \cellcolor{Gray1} NIL & \\

\hline
\hline

\textbf{Emotion} & \multicolumn2l{\textsc{CONFIDENT}} & & & & \\
\textbf{Context} & \multicolumn5l{I felt pretty good leaving the gym today, hit a new PR on overhead press}   & \\
\textbf{Reference} & \multicolumn2l{That's great. You must have been working really hard.} & & & & \\
\cline{2-7}
\makecell[r]{\colorbox{brilliantlavender}{\textsc{\modelname}}} & Awesome, that is great! Did you have a good time? I love doing that. & \cellcolor{Green1} Mid & \cellcolor{Green2} High & \cellcolor{Green1} Mid & \cellcolor{Green2} High & \multirow{5}{*}{\makecell[r]{\colorbox{Yellow1}{\textsc{MIME}}}}\\
\makecell[r]{\colorbox{ghostwhite}{\textsc{T5-GloVe}}} & \makecell[l]{That is awesome! You must be pretty proud! You should be pretty \\ proud of yourself.} & \cellcolor{Green1} Mid & \cellcolor{Green1} Mid & \cellcolor{Green1} Mid & \cellcolor{Gray1} NIL & \\
\makecell[r]{\colorbox{Yellow1}{\textsc{MIME}}} & I bet that is a lot of hard work. & \cellcolor{Green2} High & \cellcolor{Green1} Mid & \cellcolor{Green2} High & \cellcolor{Gray1} NIL & \\
\makecell[r]{\colorbox{Blue1}{\textsc{EmpDG}}} & Oh no! I am so sorry! I bet you were mad! & \cellcolor{Gray1} NIL & \cellcolor{Gray1} NIL & \cellcolor{Gray1} NIL & \cellcolor{Gray1} NIL & \\

\bottomrule
\end{tabular}
}
\endgroup
\end{table*}

As explained in \cref{sec:synthetic}, we obtained synthetic labels for the empathy-specific attributes and sentiment polarity. Based on this, we formulated our automatic evaluation of the generated responses with respect to these attributes. In particular, we feed the generated responses of each model to the pre-trained classifiers of these attributes to obtain the labels 
(\cref{sec:synthetic}). To obtain the labels for sentiment polarity, we use VADER. The 
outcome was then compared against the synthetic labels of the gold standard data. \cref{tab:synthetic_results} presents these results which suggest a noteworthy performance improvement by \modelname{} for all the attributes including sentiment polarity of the generated responses. The significant boost in performance can be seen for exploration which positively correlates with the empathy-specific human evaluation results as depicted in \cref{tab:gen_results}.

To further verify the findings of human and automatic evaluation of \cref{tab:gen_results} and \cref{tab:synthetic_results}, we have carried out the A/B testing where annotators were asked to directly compare the responses by two models and declare a win, loss, or a tie. As depicted in \cref{tab:ab}, \modelname{} outperforms SOTA and all the baselines by a large margin.

\begin{table}[ht!]
\small
    \centering
    \caption{\footnotesize Automated synthetic label-based evaluation of models; the F1 measures are all macro F1 and $mae$ stands for mean absolute error.
   }
    \newcommand\mystack[2]{$\stackrel{\mathrm{\displaystyle #1}}{(#2)}$}
    \resizebox{0.8\linewidth}{!}{
    \begin{tabular}{|l|cccc|}
    \hline
        \textbf{Method} & \makecell[c]{\textbf{Emotion} \\ \textbf{Presence}} & \makecell[c]{\textbf{Inter-}\\\textbf{pretation}} & \makecell[c]{\textbf{Explor}-\\\textbf{ation}} & \makecell[c]{\textbf{Senti-}\\\textbf{ment}}\\
        & (F1) & (F1) & (F1) & (mae) \\
    \hline
        MIME & 51.74 & 50.51 & 56.47 & 0.4484\\
        EmpDG & 55.29 & 52.58 & 52.4 & 0.4526\\
        \makecell[l]{\modelname{} (T5-GloVe) \\ \quad w/o Exemplars, \\ \quad w/o Emp. Losses} & 59.05 & 52.27 & 61 & 0.464\\
        \makecell[l]{\modelname{} \\
            \quad w/o Emp. Losses} & 59.74 & 53.44 & 61.33 & 0.453\\
        \modelname{}  & \bf 61.67 & \bf 55.55 & \bf 65.08 & \bf 0.442\\
    \hline
    \end{tabular}
    }
    \label{tab:synthetic_results}
\end{table}

\subsection{Qualitative Comparison}

We present some qualitative examples of the responses generated by the baselines and our proposed model in \Cref{tab:case_study}. On an average, we observe that \modelname{} is able to achieve `High' ratings across Relevance, Emotional Presence, and Interpolation dimensions. Even for Exploration, \modelname{} is the model that generates most amount of responses that attempt to explore the user's state further.

In the first example, we can see that \modelname's response includes not just appropriate emotional response, but the response is also inquisitive about the user's dad's health. There is also interpretation on the model's part as it tries to pacify the \textit{terrified} user by telling ``\textit{I hope it wasn't too bad}". In contrast, other baselines, like \textsc{MIME} provides apathetic response by saying ``\textit{That is good to hear}", and also insensitively concludes that the dad died. 

We also observe that in general, due to the additional losses on the empathetic dimensions, we observe \modelname's responses to be more compositional of these factors. They not just provide an appropriate empathetic phrase, but also try to interpret the situation and ask relevant questions. The baselines, in contrast, often exhibit only one of the attributes with high intensity.

Most importantly, in our qualitative exploration, we also observe one major trend regarding the association of \modelname's responses with the emotion of the user. That is, we find that on average, \modelname{} responds appropriately towards the provided emotion and does not end up being apathetic or insensitive (by being out of line). As seen in the table, examples of the user's emotion and \modelname's empathetic phrase, such as  \textit{Terrified}$\rightarrow$``\textit{Oh no}", \textit{Confident}$\rightarrow$``\textit{That is great}", \textit{Sentimental}$\rightarrow$``\textit{I'm sorry to hear that}",  \textit{Annoyed}$\rightarrow$`\textit{That sucks}", etc., all map appropriately. We believe this accuracy towards the appropriate empathetic phrase is due to the guidance provided by the retrieved exemplars. As the baseline models do not have such guidance, we observe tendencies to respond inappropriately towards a particular emotion. 

\subsubsection{\textbf{Impact of DPR}} \label{sec:DPR_impact}
We analyzed the effect of exemplars retrieved from DPR in \cref{tab:dpr_analysis} and found that it can help in two~%
ways:

1) It can pick up highly semantically similar responses that can aid the decoder to pickup correct/relevant word in the response. This is illustrated in the first example of \cref{tab:dpr_analysis}. Words generated by \modelname{} such as `confident' can be found in the exemplar as well. The overall theme of the generated response is also semantically very close to that of the exemplar. In contrast, the \emph{w/o EXEMPLAR} generation is repetitive with respect to the context, and would have been a more appropriate response in an earlier turn of the dialogue. The phrase `That is awesome' also contradicts with the speaker's feeling of being nervous.

2) It picks up responses that have stylistic similarity to the reference and pragmatically/stylistically appropriate to the context. We refer to the second example in \cref{tab:dpr_analysis}. `That's so nice of him to do.' in the exemplar is stylistically very close to `That was very kind of them.' in the reference. At the same time, the exemplar also fits stylistically with the context. The \modelname{} generation also starts with a very similar sentence as the exemplar, indicating stylistic matching. It then generates `I'm glad that you had a very good neighbor.', which displays explicit emotional presence. Overall, the \modelname{} generation can be considered to be a perfect empathetic response in this example.

\begin{table}[ht!]
\small
    \centering
    \caption{\footnotesize A/B testing results.
    }
    \resizebox{0.8\linewidth}{!}{
    \begin{tabular}{|l|l|ccc|}
        \hline
        \textbf{Model A} & \textbf{Model B} & \textbf{WIN (\%)} & \textbf{LOSS (\%)} & \textbf{TIE (\%)} \\
        \hline
        \multirow{6}{*}{\modelname{} \emph{vs.}} & MIME  & \bf 58.12  & 24.33 & 17.55 \\
        & EmpDG & \bf 56.79 & 22.50 & 20.71 \\
        & \makecell[l]{\modelname{} (T5-GloVe) \\
            \quad w/o Exemplars, \\  \quad w/o Emp. Losses}  & \bf 55.27 & 25.33 & 19.40 \\
        & T5-PT & \bf 43.13 & 31.37 & 25.50\\
        \hline
    \end{tabular}
    }
    \label{tab:ab}
\end{table}

\begin{table}[t!]
\centering
\begingroup
\setlength{\tabcolsep}{6.5pt} 
\renewcommand{\arraystretch}{1.1} 
\caption{\footnotesize Comparison of responses with and without DPR exemplars.\vspace{-1ex}}
\label{tab:dpr_analysis}
\resizebox{\linewidth}{!}
{\begin{tabular}{|l|l|}

\hline
\textbf{Emotion} & \textsc{ANXIOUS} \\
\textbf{Context} & \makecell[l]{I'm starting a new job next week, and while I am excited, I am also \\ \hspace{5ex} very nervous! \\ Congratulations!  Don't be too nervous!  You must have done \\ \hspace{5ex} something right to get the job in the first place! \\ That's true, but the uncertainty always makes me feel a bit nervous.} \\
\textbf{Reference} & That is natural.  I'm sure you will do great!\\
\textbf{Exemplar} & Just take deep breaths and act confidently and you'll do great! \\
\makecell[r]{\colorbox{brilliantlavender}{\textsc{\modelname}}} & I hope that you do try to stay confident. You will be fine. \\
\makecell[r]{\colorbox{Blue1}{\textsc{w/o Exemplar}}} & That is awesome! I wish you the best! Hope you get your job! \\

\hline
\textbf{Emotion} & \textsc{GRATEFUL} \\
\textbf{Context} & \makecell[l]{My neighbor really helped me out recently after a big windstorm.} \\
\textbf{Reference} & That was very kind of them. Did they charge you for their labor?\\
\textbf{Exemplar} & \makecell[l]{That's so nice of him to do. No one wants to help anyone out \\ \hspace{5ex} these days but I'm glad for you.} \\
\makecell[r]{\colorbox{brilliantlavender}{\textsc{\modelname}}} & \makecell[l]{That was really nice of them. I'm glad that you had a very good \\ \hspace{5ex} neighbor.} \\
\makecell[r]{\colorbox{Blue1}{\textsc{w/o Exemplar}}} & Wow that is very scary, what is the one. My family did that all the \\

\hline
\textbf{Emotion} & \textsc{FURIOUS} \\
\textbf{Context} & \makecell[l]{Not too long ago, I took a trip on a plane. The airline lost my \\ \hspace{5ex} whole suitcase!} \\
\textbf{Reference} & Oh man! how long did it take to get your stuff back?\\
\textbf{Exemplar} & Oh no! Did they reimburse you? \\
\makecell[r]{\colorbox{brilliantlavender}{\textsc{\modelname}}} & Did you call the police? They are so careless people. \\
\makecell[r]{\colorbox{Blue1}{\textsc{w/o Exemplar}}} & Oh man! That is terrible! No one happened to me in a car! \\

\hline
\textbf{Emotion} & \textsc{GRATEFUL} \\
\textbf{Context} & \makecell[l]{My dad called me one day crying out for help. I had to drive \\ \hspace{5ex} over to his house.} \\
\textbf{Reference} & Oh my goodness! What happened?\\
\textbf{Exemplar} & Why what happened? \\
\makecell[r]{\colorbox{brilliantlavender}{\textsc{\modelname}}} & Oh that sounds scary! Did he say what was going on? Is he alright? \\
\makecell[r]{\colorbox{Blue1}{\textsc{w/o Exemplar}}} & That's awful. Is he okay? Does it have someone you can do that. \\

\bottomrule
\end{tabular}
}
\endgroup
\end{table}

We find similar patterns in the other two examples in \cref{tab:dpr_analysis} as well. Exemplars provide key stylistic and semantic information which are elemental for appropriate empathetic response generation. The model without exemplars struggles to find such patterns and  finds it challenging to generate proper empathetic responses.


We further analyze the effect of DPR exemplars in \cref{tab:exemplars}. Here, we illustrate the contrasting nature of exemplars retrieved by two DPR models, where only one has been fine-tuned on the \texttt{EmpatheticDialogues} dataset. The other model is the original DPR checkpoint \cite{karpukhin-etal-2020-dense}, which has not been fine-tuned any further. We denote this models as the fine-tuned DPR and the pre-trained DPR, respectively.

In \cref{tab:exemplars}, we notice some fundamental differences in the nature of the retrieved exemplars. Fine-tuned DPR consistently provides longer and more contextually relevant exemplars. The aspect of interpretation and specific exploration can also be noticed more consistently across the exemplars. Giving a compliment--`Whoa, that is really impressive', along with asking a specific question--`How much do you bench now?' suggest contextual relevance, and the presence of the finer-grained attributes of empathy. However, exemplars from the pre-trained DPR are often much shorter and generic in nature. Very short responses such as `Why?' or `Ohh nice!' might be considered as an indication of lack of interest and non-empathetic by the listener. 

Overall, we find of trend of fine-tuned DPR providing more diverse and empathetic exemplars to be consistent across the entire dataset. The fine-tuning on \texttt{EmpatheticDialogues} ensures semantically more relevant exemplars in multi-turn contexts as well. We randomly sampled 100 instances of (context, response) pairs and found that fine-tuned DPR exemplars are better than pre-trained DPR exemplars in 92 cases.

\subsubsection{\textbf{Impact of Empathetic Losses}}
As discussed above, empathetic losses have a notable contribution to \emph{\modelname's} performance. We illustrate this effect in \cref{tab:emp_loss_analysis}. In our analysis, our human evaluators have identified several reasons behind this performance improvement:

\begin{table}[t!]
\centering
\begingroup
\setlength{\tabcolsep}{6.5pt} 
\renewcommand{\arraystretch}{1.1} 
\caption{\footnotesize Comparison of responses with and without empathetic losses.\vspace{-1ex}}
\label{tab:emp_loss_analysis}
\resizebox{\linewidth}{!}
{\begin{tabular}{|l|l|}


\hline
\textbf{Emotion} & \textsc{JOYFUL} \\
\textbf{Context} & \makecell[l]{I was so excited when I got my new puppy, it was awesome!} \\
\textbf{Reference} & \makecell[l]{Man, I'm so jealous of you. I love that new puppy stage, as tumultuous \\ \hspace{5ex} as it can be. There's just so much emotion and "new feelings" \\ \hspace{5ex} going on as you watch it grow.}\\
\makecell[r]{\colorbox{brilliantlavender}{\textsc{\modelname}}} & That's so cute! What kind of puppy is it?\\
\makecell[r]{\colorbox{Blue1}{\textsc{w/o Emp. Losses.}}} & That is so amazing, I love cats, and they are great! Love dogs! \\

\hline
\textbf{Emotion} & \textsc{SURPRISE} \\
\textbf{Context} & \makecell[l]{My new manager gave me a pay-rise!} \\
\textbf{Reference} & \makecell[l]{That's got to be an amazing feeling to have going for you. \\ \hspace{5ex} How do you feel?}\\
\makecell[r]{\colorbox{brilliantlavender}{\textsc{\modelname}}} & That's a very lucky surprise. That would be a lot of fun to have some. \\
\makecell[r]{\colorbox{Blue1}{\textsc{w/o Emp. Losses.}}} & That's good! I am sure it did.\\

\bottomrule
\end{tabular}
}
\endgroup
\end{table}

\noindent 
\begin{enumerate}
\item
After adding the empathy-specific losses, \modelname{} attains the ability to ask relevant questions to the user, and this attributes to the conspicuous performance improvement by \modelname{} for the exploration. The first example in \cref{tab:emp_loss_analysis} shows one such case where a specific question about the `puppy' is asked.
\item
Due to the inclusion of \emph{emotional presence} loss in our model, the generated responses bear more accurate and reasonable amount of emotions. In the second example in \cref{tab:emp_loss_analysis}, \modelname{} generates a response having words `surprise', `lot of fun' which showcases accurate \emph{emotional presence} with respect to the context.
\item
Exemplars play a major role in establishing the theme of the empathetic response, but it still suffers from producing generic responses. Generic responses fit contexts at the cost of poor relevance. Due to the \emph{interpretation} and \emph{exploration} losses, \modelname{} generates more relevant, contextual, and on-topic words in the responses. Consequently, this results in overall better empathetic responses having comparatively higher interpretation and exploration scores (see \cref{tab:gen_results}).
\end{enumerate}

\section{Conclusion}

We show the efficacy of employing exemplary responses from the training set in guiding the decoder to generate stylistically and thematically apt responses that communicate empathy in the context. 
Next,
we incorporate
four empathy-inducing attributes---\emph{emotional presence}, \emph{interpretation}, \emph{exploration}, and \emph{sentiment}---%
in a controlled fashion, using synthetic labels, to regulate the responses. Finally, \emph{emotional presence}, \emph{interpretation}, and \emph{exploration} attributes are also manually evaluated to better gauge the empathetic qualities of the generated responses. All these approaches collectively result in an improved framework for empathetic response generation research.

\modelname, however, does not employ an end-to-end retrieval and generation approach. The infusion of exemplars into the generative model is also rather simple and na{\"\i}ve. These reveal a number of drawbacks of the model as follows:

\begin{enumerate}
\item
\emph{Lack of relevance measurement:} \modelname{} relies on DPR for exemplar retrieval which prohibits the relative relevance ranking of the retrieved responses by the generative model with a generative cross-entropy objective. This can make the model sensitive to noise as it has no means to particularly attend to the relevant information present in the exemplars or discard any noise present in the exemplars. One could use a unique training objective, attentions on the exemplars, employ non-trivial more sound fusion techniques that can directly influence decoded output to attain better results.
\item
\emph{Worse performance for multiturn context:} We and our annotators observed that the trained DPR does not perform very well on multiturn context and often returns identical responses to the first utterance in a dialogue. This again calls for an end-to-end retrieval and generation.
\end{enumerate}



\bibliographystyle{acmart}
\bibliography{bibliography, rada, extras}
\end{document}